\documentclass{article}

\usepackage{arxiv}

\usepackage[utf8]{inputenc} 
\usepackage[T1]{fontenc}    
\usepackage{hyperref}       
\usepackage{url}            
\usepackage{booktabs}       
\usepackage{amsfonts}       
\usepackage{amsmath}
\usepackage{nicefrac}       
\usepackage{microtype}      
\usepackage{lipsum}
\usepackage{graphicx}
\usepackage[ruled]{algorithm2e} 
\graphicspath{ {./images/} }

\newtheorem{theo}{Theorem}[section]

\newtheorem{defin}{Definition}[section]

\title{A Hybrid Virtual Element Method and Deep Learning Approach for Solving One-Dimensional Euler-Bernoulli Beams}

\author{
 Paulo Akira F. Enabe \\
    Escola Politénica\\
    University of São Paulo\\
    Department of Structural and Geotechnical Engineering\\
  \texttt{paulo.enabe@usp.br} \\
   \And
 Rodrigo Provasi \\
    Escola Politénica\\
    University of São Paulo\\
    Department of Structural and Geotechnical Engineering\\
  \texttt{provasi@usp.br} \\
}

\begin{document}
\maketitle
\begin{abstract}
A hybrid framework integrating the Virtual Element Method (VEM) with deep learning is presented as an initial step toward developing efficient and flexible numerical models for one-dimensional Euler-Bernoulli beams. The primary aim is to explore a data-driven surrogate model capable of predicting displacement fields across varying material and geometric parameters while maintaining computational efficiency. Building upon VEM’s ability to handle higher-order polynomials and non-conforming discretizations, the method offers a robust numerical foundation for structural mechanics. A neural network architecture is introduced to separately process nodal and material-specific data, effectively capturing complex interactions with minimal reliance on large datasets. To address challenges in training, the model incorporates Sobolev training and GradNorm techniques, ensuring balanced loss contributions and enhanced generalization. While this framework is in its early stages, it demonstrates the potential for further refinement and development into a scalable alternative to traditional methods. The proposed approach lays the groundwork for advancing numerical and data-driven techniques in beam modeling, offering a foundation for future research in structural mechanics.
\end{abstract}


\section{Introduction}

The objective of this work is to develop and investigate a hybrid numerical framework that integrates the Virtual Element Method (VEM) with deep learning to address one-dimensional Euler-Bernoulli beam problems. The aim is to construct a surrogate model capable of predicting displacement fields efficiently across varying material and geometric configurations. By leveraging the flexibility and robustness of VEM for handling higher-order formulations and non-conforming discretizations, combined with the data-driven capabilities of deep learning, the proposed approach seeks to explore new possibilities for computational mechanics. This framework is intended as a foundation for further research, providing insights into how traditional numerical methods and modern machine learning techniques can be harmonized for enhanced performance in structural analysis.

The Virtual Element Method (VEM) is originally presented in \cite{beirao2013vem}. The method proposes to generalize the classical Finite Element Method (FEM) in terms of discretization. While FEM relies on traditional elements like triangles or quadrilaterals in two dimensions (see \cite{beirao2015elastic}, \cite{vacca2017hyperbolic}, \cite{artioli2020curvilinear}, and \cite{wriggers2020general}) and tetrahedra or hexahedra in three dimensions (see \cite{hudobivnik2019plasticity}, \cite{cihan2022contact} and \cite{xu2024highorder}), VEM is designed to work with arbitrary polygonal or polyhedral elements. This flexibility allows VEM to handle complex geometries that may not be easily meshed with standard FEM elements, such as those with non-convex shapes, highly irregular boundaries, or mixed element types within a single discretization.

The advantage of using the VEM instead of the FEM for one-dimensional beams lies primarily in its ability to simplify the formulation of higher-order elements while maintaining a consistent structure that integrates seamlessly with existing FEM frameworks. In VEM, higher-order beam and truss elements can be formulated with relative ease. These elements include additional internal variables (such as moments), which are not directly associated with nodes but contribute to capturing more complex behaviors within the element. This allows VEM to approximate higher-order deflections and stress distributions without increasing the number of nodal unknowns. For instance, in the case of a beam element, higher-order polynomials can be projected onto a reduced polynomial space, enabling a balance between computational efficiency and solution accuracy.

An important technical benefit of VEM is its reliance on projection operators that simplify the integration of internal variables, even for higher-order elements. These projection operators ensure that the method maintains accuracy and stability, making it well-suited for problems where exact polynomial solutions are desired. Additionally, VEM elements can yield exact solutions for problems where the analytical solution matches the polynomial approximation of the element, regardless of the order used.

Deep learning has emerged as a transformative approach in computational mechanics, leveraging the powerful approximation capabilities of neural networks to address complex problems in forward and inverse modeling. Rooted in the use of artificial neural networks (ANNs), deep learning excels at discovering intricate patterns in data, making it well-suited for approximating nonlinear relationships and solving differential equations. The approximation capabilities of neural networks are extensively presented in \cite{cybenko1989approximation}, \cite{hornik1989universal} and \cite{hornik1991approximation}. The Universal Approximation Theorem in which deep learning is based on, is presented in \ref{ap:johnson-lindenstrauss}. By integrating physical principles directly into network architectures or loss functions, methods such as Physics-Informed Neural Networks (PINNs) have extended the applicability of deep learning to engineering and scientific domains. This paradigm eliminates the need for extensive labeled datasets, as neural networks can be trained using governing equations, initial and boundary conditions, and physical constraints. In the context of beam problems, deep learning offers an alternative to traditional numerical methods by providing a mesh-independent framework for approximating displacement fields, stress distributions, and material behaviors. The synergy of deep learning with established numerical techniques like the Virtual Element Method (VEM) enables hybrid approaches that combine the strengths of both, paving the way for efficient and accurate solutions to complex structural mechanics problems.

Using deep learning to solve differential equations is not a new approach. The authors in \cite{lagaris1998ann} present a method where artificial neural networks (ANNs) are employed to solve ordinary and partial differential equations by formulating a trial solution that satisfies boundary conditions and depends on a trainable feedforward neural network. The training minimizes a loss function based on the residuals of the differential equation, transforming the problem into an unconstrained optimization task. This approach produces closed-form differentiable solutions, requires less memory compared to traditional methods like finite elements, and generalizes well beyond training points. The method is demonstrated on various ODEs and PDEs, highlighting its accuracy, efficiency, and potential for solving complex boundary and initial value problems.

The works in \cite{raissi2017pinns1} and \cite{raissi2017pinns2} introduce Physics-Informed Neural Networks (PINNs), a deep learning framework that embeds physical laws directly into neural network training to solve and discover nonlinear partial differential equations (PDEs). The first paper focuses on using PINNs to solve PDEs in a data-driven manner. By encoding the governing equations as constraints in the loss function, PINNs can infer solutions with minimal data, bypassing the need for discretization in time and space. The method demonstrates effectiveness in solving challenging problems like Burgers’ equation and the Schrödinger equation, showcasing PINNs’ ability to handle sharp gradients and periodic boundaries while reducing reliance on extensive datasets and classical numerical methods.

Specifically, the work in \cite{raissi2017pinns2} extends this framework to the data-driven discovery of PDEs, aiming to identify unknown parameters or even the underlying equations governing observed phenomena. Using sparse, noisy, and scattered data, PINNs leverage neural network architectures and optimization techniques to reconstruct dynamics such as the Navier-Stokes equations or Korteweg-de Vries equation. By integrating Runge-Kutta schemes for discrete-time formulations, the approach ensures robustness, enabling accurate parameter identification even with large temporal gaps between observations. Together, these studies highlight PINNs’ potential as a versatile, efficient, and accurate tool for scientific computing, capable of solving high-dimensional forward problems and discovering governing laws from data. The Physics-Informed Neural Networks are further explored in works like \cite{qian2023hyperbolic}, \cite{sharma2023stiff}, \cite{penwarden2023causal} and \cite{anagnostopoulos2024attention}.

Despite their promise, the Physical-Informed Neural Network approach faces several challenges and limitations. A primary drawback is training instability, often caused by imbalanced loss terms, such as those for governing equations, boundary conditions, and initial conditions, leading to slow convergence or inaccurate results. PINNs also suffer from spectral bias, making them less effective at capturing high-frequency components or sharp gradients in solutions, such as shocks in hyperbolic PDEs. Additionally, their high computational cost, especially in high-dimensional problems or those requiring dense collocation points, limits their scalability. This is particularly problematic for stiff equations or long temporal domains in time-dependent problems. Finally, the performance of PINNs is highly sensitive to hyperparameter choices, such as network architecture and learning rate, requiring extensive tuning. The challenges regarding PINNs are highlighted in \cite{wang2022ntk} and \cite{rathore2024losslandscape}.

Different frameworks and strategies concerning deep learning and PDEs are developed as an alternative to PINNs. In the present text, the focus is on the approaches related to solid mechanics, in particular regarding the differential equations associated with the Theory of Elasticity. For example, the integration of deep learning into the finite element method has shown significant potential for advancing computational mechanics. In \cite{jung2020deep}, the authors propose “deep learned finite elements” for two-dimensional quadrilateral elements with 4 and 8 nodes, where neural networks generate strain-displacement matrices based on geometry and material properties, automating traditional finite element procedures. This approach ensures key properties like rigid body motion and constant strain fields are preserved while achieving superior accuracy and adaptability for higher-order, three-dimensional, and nonlinear problems. Similarly, the concept of a self-updated finite element (SUFE) introduced in \cite{jung2022sufe} focuses on enhancing solution accuracy without mesh refinement by iteratively updating the stiffness matrix based on deformation and optimal bending directions derived via deep learning. SUFE minimizes shear locking, even for coarse or distorted meshes, and demonstrates superior performance in benchmark problems, highlighting its efficiency and robustness for engineering applications. Together, these methods showcase how deep learning can transform finite element modeling by improving accuracy, convergence, and computational efficiency.

The paper \cite{nguyen2020dem} introduces the Deep Energy Method (DEM), a deep learning framework that replaces traditional Finite Element Method (FEM) discretization by minimizing the system’s potential energy as a loss function. By leveraging neural network-based trial functions that inherently satisfy boundary conditions, DEM ensures equilibrium and offers a meshfree, scalable solution to nonlinear hyperelasticity problems. Evaluated on benchmark cases, including one-dimensional bars, two-dimensional bending beams, and three-dimensional torsion problems, DEM demonstrates accuracy comparable to FEM while simplifying implementation and excelling in handling complex geometries and high-dimensional datasets. Despite its promise, challenges in boundary condition imposition and numerical integration highlight opportunities for future refinement. An extension of this work is presented in \cite{abueidda2022dem}.

The work in \cite{meethal2023femnn} presents a hybrid framework combining the FEM with neural networks to enhance the efficiency and accuracy of computational mechanics. By embedding FEM’s physics-based constraints into the neural network’s loss function, the approach minimizes the need for extensive labeled datasets and ensures solutions remain consistent with physical laws. It predicts parametric PDE solutions for forward problems and identifies unknown parameters for inverse problems, with applications in uncertainty quantification, structural analysis, and rotordynamic systems. The method outperforms traditional FEM and Physics-Informed Neural Networks (PINNs) in accuracy, stability, and convergence, offering a robust and efficient alternative for solving complex engineering problems.

This work builds on the growing body of research exploring deep learning as a transformative tool in computational mechanics, particularly in the development of efficient surrogate models for structural analysis. Methods such as deep learned finite elements \cite{jung2020deep} and self-updated finite elements (SUFE) \cite{jung2022sufe} demonstrate how neural networks can automate or adapt finite element formulations to enhance accuracy and efficiency without refining meshes. Similarly, the Deep Energy Method (DEM) \cite{nguyen2020dem, abueidda2022dem} uses energy minimization principles to provide meshfree solutions for nonlinear problems, offering robust alternatives to classical FEM. Inspired by these advances, this work introduces a novel surrogate model for predicting displacement fields in structures, designed to generalize effectively across varying material properties and boundary conditions. The model incorporates a unique architecture that separates node-specific and material-specific inputs into distinct sub-networks, capturing complex interactions with improved accuracy. Additionally, the integration of Sobolev training and GradNorm dynamically balances loss components, enabling robust training and strong generalization even with limited data. This approach advances the state of surrogate modeling, providing faster and more adaptable solutions for structural mechanics compared to traditional numerical methods.

The main contributions of this work are:
\begin{itemize}
    \item \textbf{Efficient Surrogate Model for Structural Mechanics:} Developed a surrogate model for predicting displacement fields in structures that generalizes well across varying material properties and boundary conditions, enabling faster simulations compared to traditional numerical methods like FEM or VEM.
    \item \textbf{Novel Architecture Division:} The paper introduces a network architecture that separates input processing into node-specific and material-specific sub-networks, improving the representation of complex interactions between node data and material properties in structural mechanics.
    \item \textbf{Enhanced Generalization with Limited Data:} The model demonstrates strong generalization, maintaining accuracy even with fewer training samples. This characteristic increases its applicability to diverse real-world scenarios, where data availability may be constrained.
    \item \textbf{Integration of Sobolev Training and GradNorm for Balanced Losses:} Combined Sobolev training with GradNorm to dynamically balance various loss components, including data loss and material penalties. This approach ensures robust convergence and balanced learning across different aspects of the problem, enabling the model to generalize effectively even with limited training data.
\end{itemize}

\section{General-order virtual element formulation for Euler-Bernoulli beams}

In this section, the VEM formulation is presented. First, a brief introduction to the formulation of one-dimensional Euler-Bernoulli beams is shown. In the next subsection, the fundamental components for the construction of the method are described. Finally, the proper formulation of the method is presented.

The VEM offers distinct advantages for modeling one-dimensional Euler–Bernoulli beams, making it a powerful alternative to traditional numerical methods. VEM’s flexibility in handling higher-order formulations allows for the accurate representation of beam deflections and stress distributions without increasing the number of nodal unknowns. By introducing internal variables that capture higher-order effects, VEM enables the formulation of beam elements that can approximate complex deformation behaviors with ease. This is particularly beneficial for problems requiring higher-order polynomials, as VEM projects these onto a reduced polynomial space, ensuring computational efficiency while maintaining solution accuracy.

Another significant advantage of VEM is its compatibility with irregular discretizations. Unlike traditional finite elements, which may require structured meshes or specific element shapes, VEM works seamlessly with arbitrary segmentations, enabling the analysis of beams with complex geometries or boundary conditions. Furthermore, its reliance on projection operators simplifies the integration of internal variables, ensuring stability and precision even for higher-order elements. The approach presented here builds upon the works of \cite{beirao2013vem} and \cite{wriggers2022vem}, further extending VEM’s capability for efficiently modeling Euler–Bernoulli beams with general-order formulations.

The differential equation regarding the Euler-Bernoulli one-dimensional model is given by:
\begin{equation}\label{eq:euler_bernoulli}
    EI \frac{d^4 w}{d x^4}(x) = q(x),
\end{equation}
where $E$ is the elasticity modulus, $I$ is the inertia moment, $w$ is the deflection field and $q$ is the load function. The energy functional can be defined as:
\begin{equation}\label{eq:energy_functional}
    U(w) = U_K (w) - U_f (w),
\end{equation}
where
\begin{equation}\label{eq:functional_stiff}
    U_K (w) = \frac{1}{2} \int \limits^L_0 EI \left[ \frac{d^2 w}{d x^2}(x) \right]^2 dx
\end{equation}
and
\begin{equation}\label{eq:functional_load}
    U_f (w) = \int \limits^L_0 q(x) w(x) dx,
\end{equation}
with $L$ the beam's length. In this way, the idea is to find an approximated deflection field to minimize the energy functional in \ref{eq:energy_functional}.

\subsection{Virtual element formulation components}

The virtual element space is defined as a subset of a Sobolev space that includes polynomials up to a certain degree along with additional non-polynomial virtual functions. Specifically, these functions satisfy the governing equations and are determined by degrees of freedom without requiring explicit representation within the element. This space is constructed to ensure polynomial consistency, allowing the stiffness matrix to be computed exactly when interacting with polynomial test functions. The non-polynomial components, while not explicitly represented, are managed through projection operators and stabilization terms, ensuring stability and convergence of the numerical method. In this work, consider $C^0(\mathbb{F})$ represents the space of continuous functions over the field $\mathbb{F}$ and $\mathbb{P}_k(\mathbb{F})$ is the space of polynomials of degree at most $k$ on a field $\mathbb{F}$.Considering an element $e$ in an arbitrary discretization $\mathcal{T}$, the local virtual element space is:
\begin{equation}
    V_{h,e}(e) = \left\{ w_h \in H^1(e): w_h|_e \in C^0(e), \quad \Pi^\nabla w_h \in \mathbb{P}_k (e) \right\},
\end{equation}
where $H^1(e)$ is the Sobolev space, and $\Pi^\nabla w_h$ is the projection of the deflection in a polynomial space.

In the general virtual element formulation, the projection operator is used to handle terms that are non-polynomial or are polynomials of high-order. Thus, those functions are projected from the virtual element space to a low-order polynomial space. The projection operator is given by
\begin{equation}
    \Pi^\nabla: V_{h,e}(e) \longrightarrow \mathbb{P}_k (e).
\end{equation}
With the operator definition, the following three properties are enforced to guarantee the consistency of the approximation field:
\begin{equation}\label{eq:first_property}
    \int \limits^{L_e}_0 \frac{d^2 p}{dx^2} \left(\frac{d^2\Pi^\nabla w_h}{dx^2} - \frac{d^2w_h}{dx^2} \right)dx = 0,
\end{equation}

\begin{equation}\label{eq:second_property}
    \int \limits^{L_e}_0 \frac{dp}{dx} \left(\frac{d\Pi^\nabla w_h}{dx} - \frac{dw_h}{dx} \right)dx = 0
\end{equation}
and
\begin{equation}\label{eq:third_property}
    \int \limits^{L_e}_0 p \left(\Pi^\nabla w_h - w_h \right)dx = 0,
\end{equation}
where $L_e$ is the length of element $e$ and $p$ is a polynomial with the same degree as $\Pi^\nabla w_h$. The first equation given in (\ref{eq:first_property}) ensures that $\Pi^\nabla w_h$ reproduces the behavior of $w_h$ in terms of its curvature when tested against the polynomial space, preserving the second-derivative characteristics critical for capturing bending effects in beam formulations.  Equation (\ref{eq:second_property}) ensures that $\Pi^\nabla w_h$ reproduces the behavior of $w_h$ in terms of its gradient when tested against the polynomial space. And, equation \ref{eq:third_property} enforces orthogonality in the displacement space, ensuring that  $\Pi^\nabla w_h$ preserves the integral properties of $w_h$ over the element.

\subsection{Virtual element formulation}
\begin{figure}[htb] 
    \centering
    \includegraphics[width=8cm]{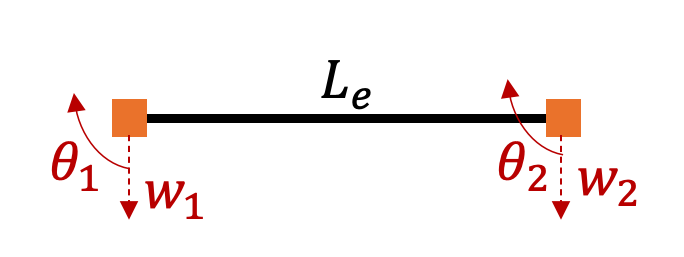}
    \caption{Sample figure caption.}
  \label{fig:element}
\end{figure}

The approximation of the deflection field is only known at the nodal points of the element. For the fourth-order differential equation presented in (\ref{eq:euler_bernoulli}), not only the deflection field but also the rotations have to be approximated. An element of the degrees of freedom is shown in Figure \ref{fig:element}. To the projected polynomial, the following choice is made:
\begin{equation}\label{eq:projection_polynomial}
    \Pi^\nabla w_h  \sum \limits^n_{k=0} a_{k+1}x^k,
\end{equation}
for $n \geq 3$. Differentiating this polynomial twice with respect to $x$:
\begin{equation}\label{eq:second_diff_proj}
    \frac{d^2 \Pi^\nabla w}{d x^2} = \underline{B}(x)\hat{a},
\end{equation}
where
\begin{equation}
    \underline{B}(x) = \left[ \begin{array}{cccc}
         2 & 6 & \cdots & k(k-1)x^{k-2}  \\ 
    \end{array} \right]
\end{equation}
and
\begin{equation}
    \hat{a} = \left[ \begin{array}{cccc}
         a_3 & a_4 & \cdots & a_{k+1}  \\ 
    \end{array} \right]^T.
\end{equation}

Considering the curvature, equation \ref{eq:first_property} can be written as:
\begin{equation}\label{eq:aux_curvature}
    \int \limits^{L_e}_0 \frac{d^2 p}{dx^2} \frac{d^2\Pi^\nabla w_h}{dx^2}dx = \int \limits^{L_e}_0 \frac{d^2 p}{dx^2} \frac{d^2w_h}{dx^2} dx.
\end{equation}
Using equation (\ref{eq:second_diff_proj}) and recalling that $p$ is a polynomial with the same degree as the polynomial projection, the left-hand side of equation (\ref{eq:aux_curvature}) can be written as:
\begin{equation}\label{eq:left_hand_curvature}
    \int \limits^{L_e}_0 \frac{d^2 p}{dx^2} \frac{d^2\Pi^\nabla w_h}{dx^2}dx = \int \limits^{L_e}_0 \underline{B}^T(x) \underline{B}(x) dx \hat{a} = \underline{G} \hat{a},
\end{equation}
with
\begin{equation}
    G = \int \limits^{L_e}_0 \underline{B}^T(x) \underline{B}(x) dx.
\end{equation}
The values of the approximation $w_h$ are not known inside the element only at the nodes, the integral by parts is applied twice on the right-hand side of equation (\ref{eq:aux_curvature}):
\begin{equation}\label{eq:right_hand_curvature}
    r = \int \limits^{L_e}_0 \frac{d^2 p}{dx^2} \frac{d^2w_h}{dx^2} dx = \left. \left( \frac{d^2p}{dx^2} \frac{d w_h}{dx} \right)\right|^{L_e}_0 - \left. \left( \frac{d^3 p}{d x^3} w_h \right) \right|^{L_e}_{0} + \int \limits^{L_e}_0 \frac{d^4 p}{d x^4}w_h dx.
\end{equation}
At the nodes, both deflection and rotation are assumed to be known:
\begin{equation}
    w_h(0) = w_1, \quad w_h(L_e) = w_2, \quad \frac{dw_h}{dx}(0) = \theta_1, \quad and \quad \frac{dw_h}{dx} (L_e) = \theta_2.
\end{equation}
Note that, for $n = 3$, the integral term in equation (\ref{eq:right_hand_curvature}) vanishes. For $n \geq 4$, an auxiliary variables must be defined. This variable is called internal moment and is given by:
\begin{equation}\label{eq:internal_moment}
    m_{k-4} = \frac{1}{L_e^{k-3}} \int \limits^{L_e}_{0} x^{k-4} w_h dx,
\end{equation}
with $4 \leq k \leq n$. The internal moment represents the moment distribution within an element that arises due to bending. It is a key internal variable used to approximate the beam’s response, especially in formulations like the Euler–Bernoulli beam theory. Unlike nodal variables, which are explicitly defined at the beam’s boundaries or interfaces, the internal moment is treated as an internal degree of freedom within the virtual element space.

With equations (\ref{eq:left_hand_curvature}) and (\ref{eq:right_hand_curvature}), the vector $\hat{a}$ is determined:
\begin{equation}
    \hat{a} = \underline{P} \hat{w},
\end{equation}
where
\begin{equation}
    \underline{P} = \underline{G}^{-1}r
\end{equation}
and
\begin{equation}
    \hat{w} = \left[ \begin{array}{cccccc}
         w_1 & \theta_1 & w_2 & \theta_2 & m_0 \cdots & m_{k-4}  \\ 
    \end{array} \right]^T.
\end{equation}
The left constants to be found are $a_1$ and $a_2$ that can be calculated by using equations (\ref{eq:second_property}) and (\ref{eq:third_property}):
\begin{equation}
    a_1 = w_1 \quad and \quad a_2 = \theta_1.
\end{equation}

With the projection polynomial $\Pi^\nabla w_h$ fully defined, the functional presented in (\ref{eq:functional_stiff}) is written as:
\begin{equation}
    U_K (w_h) = \frac{1}{2} \hat{w}^T \underline{P}^T \underline{G} \underline{P} \hat{w}.
\end{equation}
With the discretized functional, the stiffness matrix for element $e$ can be obtained:
\begin{equation}
    \underline{K} = \frac{\partial^2 U_K}{\partial w \partial w}.
\end{equation}

Regarding the load term, for $n \geq 4$, it can be written as:
\begin{equation}
    q(x) = \sum \limits^n_{k=4} q_{k-4}x^{k-4}.
\end{equation}
Thus, the functional in (\ref{eq:functional_load}) becomes:
\begin{equation}
    U_f (w_h) = \sum \limits^n_{k=4} q_{k-4}L_e^{k-3}m_{k-4}.
\end{equation}
The load vector is given by:
\begin{equation}
    f = \frac{\partial U_f}{\partial w}.
\end{equation}

To extend the virtual element formulation for Euler–Bernoulli beams to include axial effects for portico elements, it is not necessary to reformulate the entire methodology. Similar to the approach used in the Finite Element Method (FEM), the axial component can be seamlessly incorporated by adding an additional degree of freedom representing the longitudinal displacement. Unlike the bending component, the axial direction does not require internal moments, as the deformation is linear and captured directly through the axial strain.

The axial displacement field is treated analogously to the bending displacement, using the same virtual element space and projection operators to ensure consistency. The stiffness matrix is extended by adding terms that account for the axial strain energy, computed as the derivative of the axial displacement. The coupling between axial and bending behaviors, if present, is naturally handled by the flexibility of the VEM framework. This straightforward extension leverages the existing structure of the beam formulation, requiring only the addition of axial energy contributions to the functional, without the need to redefine internal variables or the stabilization terms, thus maintaining the simplicity and computational efficiency of the method.

\section{Architecture of the neural network}
\label{sec:headings}
In this work, a neural network architecture is tailored to predict displacement fields in structural mechanics problems. This architecture is designed to capture both nodal and material-specific influences by utilizing a multi-stage learning approach, where the network is divided into separate sub-networks dedicated to nodes and material properties. Two sub-networks are proposed; the first regards the nodes in the mesh, and the second one regards the material and geometric scalar parameters.

The node sub-network is responsible for processing input features related to the nodes of the structural model. It takes into account parameters such as nodal coordinates, boundary conditions, and other geometric factors that can influence the displacement fields. This network is designed with a series of dense layers, configured to learn the relationships between nodal positions and displacements effectively.

Parallel to the node sub-network, the material sub-network focuses on learning the effects of material properties, such as Young’s modulus, cross-sectional area, and moment of inertia, which are critical to modeling structural behavior. The material sub-network consists of its own set of dense layers, specifically structured to capture the nonlinear relationships between material properties and their influence on the displacement fields.

After the node and material sub-networks process their respective inputs, the outputs are concatenated. This step integrates the learned features from both components, enabling the network to make comprehensive predictions that account for both nodal influences and material behaviors. The concatenated output is then passed through a final set of dense layers to produce the final prediction for the displacement field, making it a unified model that reflects the combined effects of nodes and material properties.

The architectural division is motivated by the need to model the distinct roles that nodal geometry and material properties play in structural mechanics problems. By splitting the network into dedicated sub-networks, the model can effectively learn specialized representations for each component before combining them. 

Let $\Omega$ be a real compact space. In this text, $f \in H^k(\Omega)$ is the target function and $H^k(\Omega)$ is the Sobolev space for any non-negative $k$. This space is equipped with an inner product induced by the canonical norm, and denoted by $\langle \cdot, \cdot \rangle$. The approximation model is denoted by $m \in H^k(\Omega)$. The assumption of the approximation of functions by neural network models is based on \cite{hornik1991approximation}. The necessary set of parameters is denoted by $\theta$. The dataset is given by $n$ samples drawn independently from a data-generating distribution, and it is denoted by $(x_i)_{i\in[1,..., n]}$. Following the notation in \cite{czarnecki2017sobolev}, the Jacobian and Hessian are denoted by:
\begin{equation}
    D^j_x f(x) = \left\{ \frac{\partial^j f(x)}{\partial x_1^{\alpha_1} \partial x_2^{\alpha_2} \cdots \partial x_n^{\alpha_n}} \; \middle| \; \sum_{i=1}^n \alpha_i = j \right\},
\end{equation}
where $\alpha = (\alpha_1, \dots, \alpha_n)$ denotes the multi-index notation.

\subsection{Sobolev training}

Sobolev training is presented in \cite{czarnecki2017sobolev} and it is a technique that incorporates derivative information into the training process of neural networks. Unlike traditional training, which minimizes the error between the predicted and actual values, Sobolev training also minimizes the error of the derivatives of the predicted outputs. This approach enforces smoothness in the model’s predictions by focusing not only on the function’s values but also on how they change across the domain. It is particularly valuable in scenarios involving physical systems governed by differential equations, as it promotes physically consistent and stable predictions.

In the context of structural mechanics, Sobolev training ensures that the neural network not only approximates displacement fields accurately but also learns the underlying gradient behavior, aligning the model closely with the governing equations of the problem. This alignment enhances the network’s ability to generalize from limited data by capturing both the solution and its derivative information, thereby yielding a more stable, smooth, and physically interpretable model. 

The cost function regarding the Sobolev training is given by:
\begin{equation}
    \mathcal{L}[s; w(s)] = \sum \limits^n_{i=1} \left[ L[m(x_i; w(s)), f(x_i)] +
    \sum \limits^K_{j=1} \mathbb{E}_{v^j} \left[ L_j \left( \langle D_x^j m(x_i; w(s)), v_j \rangle, \langle D_x^j f(x_i), v_j \rangle \right) \right] \right],
\end{equation}
where $L$ and $L_j$ are loss functions, $s$ is the training step, and $v_j$ is a random vector. The step $s$ represents the training iteration or optimization step during which the model parameters $w$ are updated. To reduce the computational cost of processing the Jacobian and Hessian matrices, the cost function is written in terms of the expected value of the projection of the model and the target function in a lower-dimensional space. 

The key idea is that random projections can preserve the essential characteristics of high-dimensional data in a lower dimensional space. This idea comes from the Johnson-Lindenstrauss Lemma (see Appendix \ref{ap:johnson-lindenstrauss}), which states that random projections can approximately preserve the distances between points when projecting from high to low dimensions. In this sense, the expectation $\mathbb{E}_{v_j}[\cdot]$ in the loss function means that, during training, the network computes the loss function over multiple random vectors. This averaging process ensures that the network learns the overall structure of the Jacobian or Hessian matrix without explicitly computing the full matrices.

It is important to mention that the random vectors $v_j$ are chosen from the unit sphere in the input or parameter space. Sampling from the unit sphere ensures that every possible direction in the input or parameter space is equally likely.

\subsection{GradNorm}

GradNorm is presented in \cite{chen2018gradnorm} and it is a loss-balancing technique designed to address the challenge of optimizing multiple loss functions simultaneously, each of which may have vastly different scales or rates of convergence. Instead of applying fixed weights to different losses, GradNorm dynamically adjusts these weights by monitoring the gradient norms associated with each loss term, ensuring balanced and effective learning across all tasks during training.

In neural network training, especially when multiple objectives are involved, imbalances can arise as different losses converge at different rates. For instance, a high-value loss might dominate, leading the network to focus on minimizing that loss at the expense of others, ultimately resulting in poor generalization. GradNorm corrects for this by scaling the gradients of each loss so that they all contribute equitably, which helps the model achieve more consistent and stable convergence. Different from classical batch normalization, GradNorm normalizes across tasks instead of across batches.

In this work, GradNorm is utilized to balance the contributions of displacement, Sobolev, and material penalty losses during training, enabling the model to efficiently learn across these interrelated components. By adapting the loss contributions in real time, GradNorm helps to avoid overfitting to any single objective while preserving the benefits of Sobolev training and other architectural elements. This balanced approach improves both convergence and the model’s robustness, enabling the network to generalize more effectively from limited data.

To normalize the objective values, a reference must be set. Commonly, the average gradient norm of all tasks at training step $s$ is used:
\begin{equation}
    \overline{G}_\theta(s) = \mathbb{E}_{task}\left[ G_\theta^{i}(s) \right],
\end{equation}
where
\begin{equation}
    G_\theta^i (s) = \| \nabla_\theta \theta_i (s) L_i[s; w(s)] \|
\end{equation}
is the $L^2$-norm of the gradient of the weighted single task objective function $\theta_i(s) L_i(s)$ with respect to $\Theta$. Here, $\Theta$ is the subset of the full network weights $\Theta \subset \mathcal{W}$, where the GradNorm is actually applied. As stated in the original paper, $\Theta$ is generally chosen as the last shared layer of weights for the method to be computationally efficient. 

The loss ratio for task  $i$ in the step $s$ is given by:
\begin{equation}
    \tilde{L}_i(s) = \frac{L_i[s; w(s)]}{L_i[0; w(s)]}. 
\end{equation}
The ratio $\tilde{L}_i$ is a measure of the inverse training rate of task $i$. In other words, lower values of $\tilde{L}$ correspond to a faster training rate for task $i$. The relative inverse training rate is given by:
\begin{equation}
    r_i(s) = \frac{\tilde{L}_i(s)}{\mathbb{E}_{task}[\tilde{L}_i(s)]}.
\end{equation}
With $r_i$ it is possible to rate balance the gradients. The higher the value of $r_i$, the higher the gradient magnitudes should be for task $i$ in order to encourage the task to train more quickly. Therefore, the gradient norm for each task $i$ is computed as:
\begin{equation}\label{eq:norm}
    G_\theta^{(i)} (s) \rightarrow \overline{G}_\theta (s) [r_i(s)]^\alpha,
\end{equation}
where $\alpha$ is a hyperparameter. As stated by the authors in \cite{chen2018gradnorm}, when tasks vary significantly in complexity, resulting in substantially different learning dynamics, a higher value of  $\alpha$  is recommended to enforce a stronger balance in training rates across tasks.

Equation (\ref{eq:norm}) defines a target gradient norm for each task  i . The loss weights are then adjusted to align the gradient norms of the tasks with their respective targets. GradNorm is represented by the following norm:
\begin{equation}\label{eq:grad_norm}
    \mathcal{L}_{grad}[s; \theta_i(s)] = \sum \limits^\tau_{i=1}\left| G_\theta^{(i)}(s) - \overline{G}_\theta (s) [r_i(s)]^\alpha \right|,
\end{equation}
where $\tau$ is the total number of tasks. It is important to mention that in equation (\ref{eq:grad_norm}), the $L_1$-norm is used. This norm ensures a stable linear penalty for deviations between the actual and target gradient norms. Specifically, linear penalties make the optimization process less sensitive to outliers or extreme deviations in gradient norms. When differentiating the objective function, the target gradient norm  $\overline{G}_\theta(s) [r_i(s)]^\alpha$  is treated as a fixed constant to prevent the loss weights from inadvertently converging to zero.

After every update step, the weights are normalized so that:
\begin{equation}
    \sum \limits^T_{i=1} \theta_i (s) = \tau.
\end{equation}
Without normalization, the task specific weights $\theta_i$ could grow indefinitely or shrink to zero, which would skew the training process. Large weights would make one task dominate the training process. The renormalization process is given by:
\begin{equation}
    \theta_i (s+1) \leftarrow \theta_i(s+1) \frac{\tau}{\sum_j \theta_j(s+1)}.
\end{equation}

\subsection{Training the neural network}

The training pipeline presented in Algorithm \ref{alg:training_gradnorm} integrates GradNorm for loss balancing and Sobolev training to improve the neural network’s performance in structural mechanics problems. GradNorm dynamically adjusts the relative importance of multiple loss terms during training, ensuring balanced learning across tasks, while Sobolev training enforces smoothness and physical consistency by incorporating derivative information into the loss function.

The algorithm begins by initializing the network parameters $\theta$ and the task-specific loss weights $\theta_i(0) = 1$ for each of the three tasks: displacement loss, Sobolev loss, and material penalty loss. Two optimizers are employed: Adam for updating the network parameters $\theta$, and SGD for updating the task-specific loss weights $\theta_i$. At each epoch $s$, the algorithm iterates over a dataset consisting of nodes $x_i$, material parameters, and corresponding displacements. First, the inputs $x_i$  (nodes and material parameters) are normalized, and the network  $m(x_i; \theta)$  predicts the displacement field through a forward pass. Three losses are computed: 
\begin{enumerate}
    \item the displacement loss, which measures the error between the predicted and reference displacements,
    \item the Sobolev loss, which accounts for the gradients of the solution to enforce smoothness, and
    \item the material penalty loss, which penalizes deviations in material properties.
\end{enumerate}
The total loss is a weighted sum of these three components:
\begin{equation}
    \mathcal{L}_\text{total} = \theta_1 \mathcal{L}_1 + \theta_2 \mathcal{L}_2 + \theta_3 \mathcal{L}_3,
\end{equation}
where  $\mathcal{L}_1$, $\mathcal{L}_2$, $\mathcal{L}_3$ correspond to the displacement loss, Sobolev loss, and material penalty loss, respectively, and  $\theta_i$  are the loss weights.

To dynamically balance the losses, GradNorm computes the gradient norms $G_\theta^{(i)}$  of each task and compares them to their respective targets, which are derived from the initial loss values. The GradNorm loss ensures that the gradients of the tasks scale proportionally, preventing any single task from dominating the learning process. The loss weights $\theta_i$ are updated using the gradients of the GradNorm loss. Once the loss weights are updated, backpropagation is performed on the total loss  $\mathcal{L}_\text{total}$ to update the network parameters $\theta$. The loss weights are then renormalized to maintain their sum at a constant value $T$, ensuring stability during training. If any network parameters exceed a predefined threshold, training is stopped to avoid numerical instabilities.

By combining GradNorm for loss balancing and Sobolev training for smoothness, the algorithm achieves more accurate and physically consistent predictions, even when trained on limited data. The dynamic adjustment of task weights and incorporation of derivative information enhance the model’s robustness and generalization capability, making it particularly suitable for problems governed by differential equations.

The proposed training approach models this problem by approximating the displacement field $m(x; \theta)$ using a neural network that incorporates both nodal input (beam geometry and loading conditions) and beam parameters (e.g., $E$, $I$). The Sobolev loss enforces smoothness and physical consistency by minimizing errors not only in the displacement field but also in its derivatives, ensuring the model respects the fourth-order nature of the Euler-Bernoulli equation. By integrating GradNorm, the algorithm balances the contributions of the displacement loss, Sobolev loss, and material penalty loss dynamically during training. This ensures that all aspects of the solution—displacement field accuracy, smooth gradient behavior, and consistency with material parameters—are learned effectively, even with limited training data.

Such an approach allows the neural network to capture the key physical behaviors of the beam, such as deflection patterns and gradients, while maintaining stability and robustness in the predictions. By leveraging both Sobolev training and GradNorm, the model aligns closely with the physics of the Euler-Bernoulli beam problem, providing a computationally efficient and physically interpretable surrogate for traditional numerical methods like the FEM.

\begin{algorithm}[h]
\caption{Training Neural Networks with Sobolev Loss and GradNorm}
\label{alg:training_gradnorm}
\KwData{$\Omega$: compact space, $(x_i)_{i=1}^n$: dataset, $f \in H^k(\Omega)$: target function, \\ 
$\theta$: network parameters, $m \in H^k(\Omega)$: approximation model, \\ 
$\alpha$: balancing coefficient for GradNorm, $T$: total weight normalization factor}
\KwResult{Trained model $m(x; \theta)$}
\textbf{Initialize} task weights $\theta_i(0) = 1 \; \forall i$ \\
\textbf{Initialize} network parameters $\theta$ \\
\textbf{Compute} initial task losses $L_i(0) \; \forall i$ \\
\For{$s = 1$ to $S$ \tcp*[h]{Number of training steps}}{
    \tcp{Forward Pass}
    \For{each $x_i \in (x_i)_{i=1}^n$}{
        Compute the prediction $m(x_i; \theta)$ \\
        Compute the displacement loss: \\
        \[
        L_1 = \| m(x_i; \theta) - f(x_i) \|^2
        \]
        Compute the Sobolev loss: \\
        \[
        L_2 = \sum_{j=1}^K \mathbb{E}_{v^j} \left[ \left\langle D_x^j m(x_i; \theta), v_j \right\rangle - \left\langle D_x^j f(x_i), v_j \right\rangle \right]^2
        \]
        Compute the material penalty loss: $L_3$ (problem-specific regularization)
    }
    
    \tcp{Loss Balancing with GradNorm}
    Compute the total weighted loss: \\
    \[
    \mathcal{L}_\text{total} = \sum_{i=1}^3 \theta_i(s) L_i
    \]
    Compute the gradient norms for each task: \\
    \[
    G_\theta^{(i)}(s) = \| \nabla_\theta \theta_i(s) L_i \|
    \]
    Compute the average gradient norm: \\
    \[
    \overline{G}_\theta(s) = \mathbb{E}_{task}[G_\theta^{(i)}(s)]
    \]
    Compute the loss ratio: \\
    \[
    r_i(s) = \frac{\tilde{L}_i(s)}{\mathbb{E}_{task}[\tilde{L}_i(s)]}, \quad \tilde{L}_i(s) = \frac{L_i}{L_i(0)}
    \]
    Compute the GradNorm loss: \\
    \[
    \mathcal{L}_\text{grad} = \sum_{i=1}^3 \left| G_\theta^{(i)}(s) - \overline{G}_\theta(s) [r_i(s)]^\alpha \right|
    \]
    
    \tcp{Gradient Updates}
    Update task weights $\theta_i$ using GradNorm gradients: \\
    \[
    \theta_i(s+1) \leftarrow \theta_i(s) - \eta_\theta \nabla_\theta \mathcal{L}_\text{grad}
    \]
    Update network parameters $\theta$ using the total loss: \\
    \[
    \theta \leftarrow \theta - \eta \nabla_\theta \mathcal{L}_\text{total}
    \]
    Renormalize the task weights: \\
    \[
    \theta_i(s+1) \leftarrow \theta_i(s+1) \frac{T}{\sum_{j=1}^3 \theta_j(s+1)}
    \]
}
\textbf{Return} trained model $m(x; \theta)$
\end{algorithm}

\section{Numerical results}

In this numerical simulation, the portico geometry, a rectangular structure composed of vertical and horizontal beams, was used to evaluate the hybrid deep learning and Virtual Element Method (VEM) approach. The geometry was discretized into finite elements along each edge, and two different models, quadratic (order 4) and cubic (order 5), were tested to assess the accuracy of the method.

The training of the hybrid model was performed using a dataset comprising 80 samples, each representing unique combinations of material and geometric properties, including elasticity modulus, cross-sectional area, and moment of inertia, alongside their corresponding VEM solutions. This dataset provided sufficient diversity for the deep learning model to effectively learn the underlying relationships between material parameters, geometry, and displacement fields. The training process incorporated advanced techniques such as GradNorm for dynamic loss balancing and Sobolev training to ensure smooth and physically consistent predictions.

Notably, the training was conducted on an arm64 processor using PyTorch with Metal Performance Shaders (MPS) for acceleration. This configuration highlights the approach's adaptability to modern hardware, demonstrating its practicality for computationally intensive tasks without requiring extensive high-performance computing resources. The successful training under these conditions underscores the method's efficiency and scalability for real-world applications.

The numerical simulation involved comparing the displacement field predictions from the deep learning model with reference solutions obtained using the VEM formulation. Specifically, the $H^1$ error was computed to quantify the difference between the deep learning inference and the VEM solution. This error metric accounts for both the magnitude of the displacement field and its gradient, making it suitable for evaluating the physical consistency and smoothness of the predicted solutions. For more details about the $H^1$ error, refer to Appendix \ref{ap:johnson-lindenstrauss}.

For each configuration, varying numbers of elements per edge were used to refine the discretization, and the corresponding mean and standard deviation of the  $H^1$  error were recorded across a test dataset of 20 samples. This setup enabled the investigation of the model’s convergence behavior, the impact of deep learning integration, and the differences in performance between the quadratic and cubic models.

\begin{figure}[!h] 
    \centering
    \includegraphics[width=8cm]{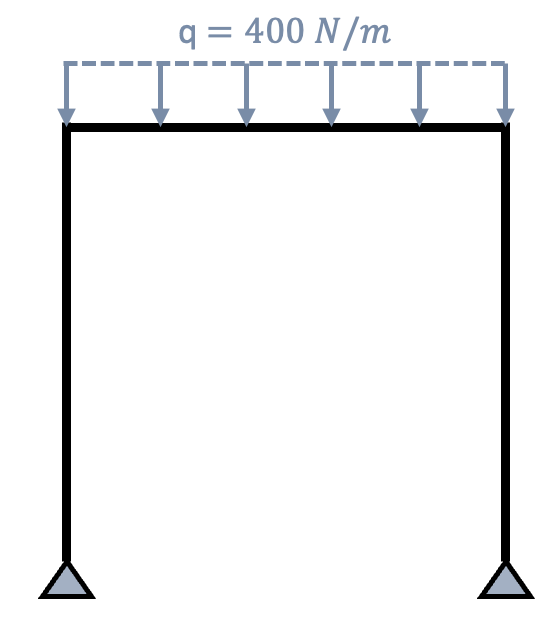}
    \caption{Portico geometry.}
  \label{fig:geometry}
\end{figure}

The portico geometry used in this study represents a classic structural system consisting of two vertical columns and a horizontal beam forming a rigid rectangular frame. The structure is supported at the base of the columns, modeled as pinned supports, which allow rotation but prevent translation in any direction. Each beam has length $L=2.0m$. The considered geometry is shown in Figure \ref{fig:geometry}.

\begin{figure}[!h] 
    \centering
    \includegraphics[width=16.5cm]{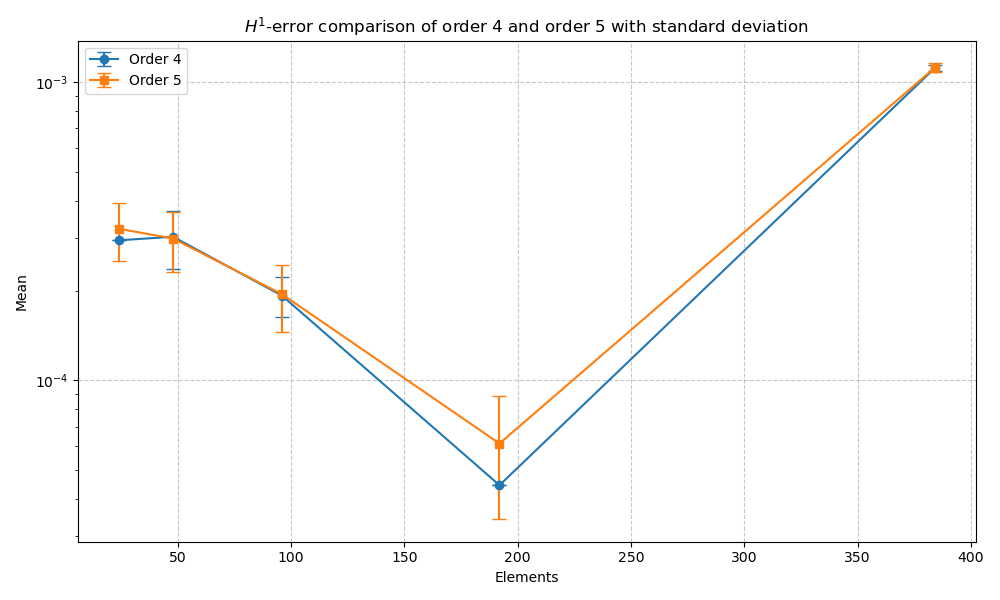}
    \caption{$H^1$-error comparison of order 4 and order 5 with standard deviation.}
  \label{fig:h1_error}
\end{figure}

The results of the hybrid deep learning and VEM approach for quadratic (order 4) and cubic (order 5) formulations, shown in Figure \ref{fig:h1_error}, provide valuable insights into the behavior of the models under different levels of discretization. For both models, the $H^1$-error mean generally decreases as the number of elements increases, demonstrating improved accuracy with finer finite element discretizations. However, a few key differences emerge between the two formulations.

The quadratic model (order 4) consistently achieves slightly lower mean errors compared to the cubic model (order 5) for most element configurations. At lower element counts (e.g., 24 and 48 elements), the errors for both models are comparable. As the resolution increases to 96 and 192 elements, the quadratic model exhibits a more pronounced reduction in error, outperforming the cubic model. Interestingly, at the highest resolution of 384 elements, both models display a noticeable increase in the H1 error mean, with similar error values. This increase could indicate challenges in the hybrid approach at very fine resolutions, potentially caused by numerical instabilities, overfitting, or limitations in the neural network’s ability to generalize accurately.

The standard deviation (Std) analysis further reveals differences in the behavior of the two models. At lower element counts, the cubic model demonstrates a higher standard deviation compared to the quadratic model, suggesting greater variability or reduced stability in its predictions. As the number of elements increases, the standard deviation decreases for both models, indicating enhanced consistency at higher resolutions. However, the cubic model maintains slightly higher variability across resolutions compared to the quadratic model.

At 384 elements, the unexpected increase in the H1 error mean for both models is particularly noteworthy. This behavior deviates from the theoretically expected convergence trends for quadratic and cubic VEM formulations. For the quadratic model, a quadratic rate of convergence is anticipated, while the cubic model is expected to exhibit higher-order convergence. The observed results suggest that the non-linear nature of the deep learning component may interfere with the ideal convergence rates. The hybrid approach introduces complexities that could disrupt the theoretical error decay associated with standard VEM formulations.

\subsection{A brief discussion about nonlinearity in deep learning}
Deep learning models are inherently non-linear approximators. Unlike the Virtual Element Method (VEM), which uses predefined polynomial spaces (quadratic or cubic) for basis functions, neural networks learn their approximation space from the training data. While this non-linear nature enables neural networks to approximate highly complex functions, it also introduces a dependence on factors such as the training process, network architecture, and data distribution. This flexibility and dependence can result in behavior that deviates from the strictly theoretical convergence expected in VEM formulations.

In the hybrid approach, the neural network is responsible for approximating components like displacements and gradients. However, the representation learned by the network may not align perfectly with the theoretical polynomial convergence rates of quadratic or cubic VEM. The interaction between the neural network’s learned approximation and the VEM formulation, which involves different mathematical assumptions, can introduce non-linearities that affect the error convergence behavior.

The training process and loss function design also play a critical role in shaping the model’s error behavior. Techniques such as Sobolev training enforce smoothness and accuracy of derivatives but do not inherently guarantee that the model will adhere to the strict polynomial error decay expected in VEM. Additionally, GradNorm dynamically adjusts the weights of different loss terms during training, which can influence convergence behavior in ways that differ from purely theoretical expectations.

A fundamental mismatch between the polynomial spaces of VEM and the learned function spaces of the neural network could also contribute to the observed deviations. VEM’s polynomial basis functions provide a predictable and controlled error convergence, whereas the neural network’s learned function space may diverge from the polynomial basis, especially if biases or artifacts are introduced by the training data, network architecture, or loss function design.

\subsection{Surrogate model}
Surrogate models serve as efficient approximations of computationally expensive numerical simulations, enabling faster predictions while retaining sufficient accuracy for practical applications. These models have gained significant attention in fields where traditional methods, such as finite element methods (FEM) or the Virtual Element Method (VEM), may require considerable computational resources. The hybrid approach presented in this work aligns closely with the concept of surrogate modeling by integrating the strengths of deep learning and numerical methods to develop a robust, data-driven framework for approximating displacement fields in structural mechanics problems.

The hybrid architecture, which combines deep learning with VEM, serves as a surrogate model by leveraging the predictive power of neural networks to emulate the behavior of numerical solutions. Specifically, the separation of node-based and material-based inputs within the architecture enables the neural network to effectively capture the relationships between geometry, material properties, and displacement fields. This approach not only accelerates predictions but also ensures that the surrogate model remains physically consistent with the underlying principles of elasticity, thanks to the incorporation of Sobolev training and GradNorm for loss balancing.

One of the key benefits of this hybrid surrogate model is its ability to generalize from limited data. In traditional surrogate modeling, generating training data often involves repeated high-fidelity numerical simulations, which can be computationally expensive. By relying on a relatively small dataset of VEM solutions, the hybrid approach demonstrates the potential to significantly reduce the data generation burden. This efficiency is achieved through advanced loss functions and training techniques that enforce smoothness, derivative accuracy, and balanced learning across tasks, ensuring that the model captures the essential physics with fewer data points.

Additionally, the practical advantages of the hybrid approach extend beyond its data efficiency. Once trained, the neural network can perform rapid inference, providing displacement field predictions in real-time or near real-time, which is particularly valuable for applications such as optimization, sensitivity analysis, and uncertainty quantification. These capabilities make the hybrid model an appealing surrogate for scenarios requiring repeated evaluations or where computational speed is critical.

\section{Conclusion}
This paper presents a promising hybrid approach that integrates Deep Learning with the Virtual Element Method (VEM) to solve structural mechanics problems, focusing on the one-dimensional Euler-Bernoulli beam and portico geometries. The method employs a neural network architecture that separately models node-based and material-based inputs, utilizing advanced techniques such as Sobolev training and GradNorm to balance losses and maintain physical consistency. The results demonstrate the approach’s practical advantages, including its ability to achieve accurate displacement field approximations with a remarkably small amount of training data and the efficiency of performing inference compared to traditional methods. These strengths underscore the potential of combining data-driven models with numerical methods for solving complex engineering problems.

The findings highlight the practical feasibility and versatility of the hybrid approach, particularly in its ability to generalize from limited data. While the observed deviations in error behavior at high resolutions raise questions about the interaction between deep learning and VEM, they also reflect the challenges of aligning non-linear neural approximations with the expected convergence behavior of traditional polynomial formulations. The unexpected similarity in performance between quadratic and cubic models further suggests opportunities for optimizing the neural network architecture and training process to fully leverage the capabilities of higher-order VEM formulations.

Future work should aim to refine this hybrid methodology by isolating the effects of deep learning components through convergence analyses using pure VEM. Enhancing the neural network’s ability to align with the physics of the problem through improved loss functions, architecture adjustments, or additional constraints could address current limitations. Furthermore, the practical implications of the method—such as efficient inference and adaptability to varying resolutions—highlight its potential for broader application in structural mechanics and beyond.

In conclusion, the integration of deep learning and VEM represents a significant step forward in bridging data-driven and physics-based methods. The demonstrated accuracy, stability, and efficiency of the approach underscore its potential to revolutionize numerical simulations in structural mechanics, offering a foundation for further innovation in hybrid computational methodologies.

\newpage

\bibliographystyle{unsrt}  


\begin{thebibliography}{1}

\bibitem{beirao2013vem}
L. Beirão da Veiga, F. Brezzi, A. Cangiani, G. Manzini, L. D. Marini, and A. Russo.
\newblock Basic principles of virtual element methods.
\newblock In {\em Mathematical Models and Methods in Applied Sciences}, Volume 23, Issue 1, pages 199--214, 2013.
\newblock \url{https://doi.org/10.1142/S0218202512500492}.

\bibitem{beirao2015elastic}
L. Beirão da Veiga, C. Lovadina, and D. Mora.
\newblock A Virtual Element Method for elastic and inelastic problems on polytope meshes.
\newblock In {\em Computer Methods in Applied Mechanics and Engineering}, Volume 295, pages 327--346, 2015.
\newblock ISSN 0045-7825.
\newblock \url{https://doi.org/10.1016/j.cma.2015.07.013}.

\bibitem{vacca2017hyperbolic}
Giuseppe Vacca.
\newblock Virtual Element Methods for hyperbolic problems on polygonal meshes.
\newblock In {\em Computers \& Mathematics with Applications}, Volume 74, Issue 5, pages 882--898, 2017.
\newblock ISSN 0898-1221.
\newblock \url{https://doi.org/10.1016/j.camwa.2016.04.029}.

\bibitem{artioli2020curvilinear}
E. Artioli, L. Beirão da Veiga, and F. Dassi.
\newblock Curvilinear Virtual Elements for 2D solid mechanics applications.
\newblock In {\em Computer Methods in Applied Mechanics and Engineering}, Volume 359, page 112667, 2020.
\newblock ISSN 0045-7825.
\newblock \url{https://doi.org/10.1016/j.cma.2019.112667}.

\bibitem{wriggers2020general}
P. Wriggers, B. Hudobivnik, and F. Aldakheel.
\newblock A virtual element formulation for general element shapes.
\newblock In {\em Computational Mechanics}, Volume 66, Number 4, pages 963--977, 2020.
\newblock \url{https://doi.org/10.1007/s00466-020-01891-5}.

\bibitem{hudobivnik2019plasticity}
Blaž Hudobivnik, Fadi Aldakheel, and Peter Wriggers.
\newblock A low order 3D virtual element formulation for finite elasto–plastic deformations.
\newblock In {\em Computational Mechanics}, Volume 63, Number 2, pages 253--269, 2019.
\newblock \url{https://doi.org/10.1007/s00466-018-1593-6}.

\bibitem{cihan2022contact}
Mertcan Cihan, Blaž Hudobivnik, Jože Korelc, and Peter Wriggers.
\newblock A virtual element method for 3D contact problems with non-conforming meshes.
\newblock In {\em Computer Methods in Applied Mechanics and Engineering}, Volume 402, page 115385, 2022.
\newblock ISSN 0045-7825.
\newblock \url{https://doi.org/10.1016/j.cma.2022.115385}.

\bibitem{xu2024highorder}
Bing-Bing Xu, Wei-Long Fan, and Peter Wriggers.
\newblock High-order 3D virtual element method for linear and nonlinear elasticity.
\newblock In {\em Computer Methods in Applied Mechanics and Engineering}, Volume 431, page 117258, 2024.
\newblock ISSN 0045-7825.
\newblock \url{https://doi.org/10.1016/j.cma.2024.117258}.

\bibitem{wriggers2022vem}
P. Wriggers.
\newblock On a virtual element formulation for trusses and beams.
\newblock In {\em Archive of Applied Mechanics}, Volume 92, pages 1655--1678, 2022.
\newblock \url{https://doi.org/10.1007/s00419-022-02113-5}.

\bibitem{cybenko1989approximation}
G. Cybenko.
\newblock Approximation by superpositions of a sigmoidal function.
\newblock In {\em Mathematics of Control, Signals, and Systems}, Volume 2, pages 303--314, 1989.
\newblock \url{https://doi.org/10.1007/BF02551274}.

\bibitem{hornik1989universal}
Kurt Hornik, Maxwell Stinchcombe, and Halbert White.
\newblock Multilayer feedforward networks are universal approximators.
\newblock In {\em Neural Networks}, Volume 2, Issue 5, pages 359--366, 1989.
\newblock ISSN 0893-6080.
\newblock \url{https://doi.org/10.1016/0893-6080(89)90020-8}.

\bibitem{hornik1991approximation}
Kurt Hornik.
\newblock Approximation capabilities of multilayer feedforward networks.
\newblock In {\em Neural Networks}, Volume 4, Issue 2, pages 251--257, 1991.
\newblock ISSN 0893-6080.
\newblock \url{https://doi.org/10.1016/0893-6080(91)90009-T}.

\bibitem{lagaris1998ann}
I. E. Lagaris, A. Likas, and D. I. Fotiadis.
\newblock Artificial neural networks for solving ordinary and partial differential equations.
\newblock In {\em IEEE Transactions on Neural Networks}, Volume 9, Number 5, pages 987--1000, September 1998.
\newblock \url{https://doi.org/10.1109/72.712178}.

\bibitem{raissi2017pinns1}
Maziar Raissi, Paris Perdikaris, and George Em Karniadakis.
\newblock Physics Informed Deep Learning (Part I): Data-driven Solutions of Nonlinear Partial Differential Equations.
\newblock Preprint, 2017.
\newblock \url{https://arxiv.org/abs/1711.10561}.

\bibitem{raissi2017pinns2}
Maziar Raissi, Paris Perdikaris, and George Em Karniadakis.
\newblock Physics Informed Deep Learning (Part II): Data-driven Discovery of Nonlinear Partial Differential Equations.
\newblock Preprint, 2017.
\newblock \url{https://arxiv.org/abs/1711.10566}.

\bibitem{qian2023hyperbolic}
Yanxia Qian, Yongchao Zhang, Yunqing Huang, and Suchuan Dong.
\newblock Physics-informed neural networks for approximating dynamic (hyperbolic) PDEs of second order in time: Error analysis and algorithms.
\newblock In {\em Journal of Computational Physics}, Volume 495, page 112527, 2023.
\newblock ISSN 0021-9991.
\newblock \url{https://doi.org/10.1016/j.jcp.2023.112527}.

\bibitem{sharma2023stiff}
P. Sharma, L. Evans, M. Tindall, and others.
\newblock Stiff-PDEs and Physics-Informed Neural Networks.
\newblock In {\em Archives of Computational Methods in Engineering}, Volume 30, pages 2929--2958, 2023.
\newblock \url{https://doi.org/10.1007/s11831-023-09890-4}.

\bibitem{penwarden2023causal}
Michael Penwarden, Ameya D. Jagtap, Shandian Zhe, George Em Karniadakis, and Robert M. Kirby.
\newblock A unified scalable framework for causal sweeping strategies for Physics-Informed Neural Networks (PINNs) and their temporal decompositions.
\newblock In {\em Journal of Computational Physics}, Volume 493, page 112464, 2023.
\newblock ISSN 0021-9991.
\newblock \url{https://doi.org/10.1016/j.jcp.2023.112464}.

\bibitem{anagnostopoulos2024attention}
Sokratis J. Anagnostopoulos, Juan Diego Toscano, Nikolaos Stergiopulos, and George Em Karniadakis.
\newblock Residual-based attention in physics-informed neural networks.
\newblock In {\em Computer Methods in Applied Mechanics and Engineering}, Volume 421, page 116805, 2024.
\newblock ISSN 0045-7825.
\newblock \url{https://doi.org/10.1016/j.cma.2024.116805}.

\bibitem{wang2022ntk}
Sifan Wang, Xinling Yu, and Paris Perdikaris.
\newblock When and why PINNs fail to train: A neural tangent kernel perspective.
\newblock In {\em Journal of Computational Physics}, Volume 449, page 110768, 2022.
\newblock ISSN 0021-9991.
\newblock \url{https://doi.org/10.1016/j.jcp.2021.110768}.

\bibitem{rathore2024losslandscape}
Pratik Rathore, Weimu Lei, Zachary Frangella, Lu Lu, and Madeleine Udell.
\newblock Challenges in Training PINNs: A Loss Landscape Perspective.
\newblock Preprint, 2024.
\newblock \url{https://arxiv.org/abs/2402.01868}.

\bibitem{jung2020deep}
Jaeho Jung, Kyungho Yoon, and Phill-Seung Lee.
\newblock Deep learned finite elements.
\newblock In {\em Computer Methods in Applied Mechanics and Engineering}, Volume 372, page 113401, 2020.
\newblock ISSN 0045-7825.
\newblock \url{https://doi.org/10.1016/j.cma.2020.113401}.

\bibitem{jung2022sufe}
Jaeho Jung, Hyunok Jun, and Phill-Seung Lee.
\newblock Self-updated four-node finite element using deep learning.
\newblock In {\em Computational Mechanics}, Volume 69, pages 23--44, 2022.
\newblock \url{https://doi.org/10.1007/s00466-021-02081-7}.

\bibitem{nguyen2020dem}
Vien Minh Nguyen-Thanh, Xiaoying Zhuang, and Timon Rabczuk.
\newblock A deep energy method for finite deformation hyperelasticity.
\newblock In {\em European Journal of Mechanics - A/Solids}, Volume 80, page 103874, 2020.
\newblock ISSN 0997-7538.
\newblock \url{https://doi.org/10.1016/j.euromechsol.2019.103874}.

\bibitem{abueidda2022dem}
Diab W. Abueidda, Seid Koric, Rashid Abu Al-Rub, Corey M. Parrott, Kai A. James, and Nahil A. Sobh.
\newblock A deep learning energy method for hyperelasticity and viscoelasticity.
\newblock In {\em European Journal of Mechanics - A/Solids}, Volume 95, page 104639, 2022.
\newblock ISSN 0997-7538.
\newblock \url{https://doi.org/10.1016/j.euromechsol.2022.104639}.

\bibitem{meethal2023femnn}
R. E. Meethal, A. Kodakkal, M. Khalil, and others.
\newblock Finite element method-enhanced neural network for forward and inverse problems.
\newblock In {\em Advances in Modeling and Simulation in Engineering Sciences}, Volume 10, page 6, 2023.
\newblock \url{https://doi.org/10.1186/s40323-023-00243-1}.

\bibitem{czarnecki2017sobolev}
Wojciech Marian Czarnecki, Simon Osindero, Max Jaderberg, Grzegorz Świrszcz, and Razvan Pascanu.
\newblock Sobolev training for neural networks.
\newblock {\em arXiv preprint arXiv:1706.04859}, 2017.
\newblock \url{https://arxiv.org/abs/1706.04859}.

\bibitem{chen2018gradnorm}
Zhao Chen, Vijay Badrinarayanan, Chen-Yu Lee, and Andrew Rabinovich.
\newblock GradNorm: Gradient normalization for adaptive loss balancing in deep multitask networks.
\newblock {\em arXiv preprint arXiv:1711.02257}, 2018.
\newblock \url{https://arxiv.org/abs/1711.02257}.

\end{thebibliography}

\newpage

\appendix

\section{Functional analysis definitions and results}
\label{ap:johnson-lindenstrauss}

The Universal Approximation Theorem states that multi-layer feedforward networks with continuous non-constant activation functions can approximate any continuous function on compact subsets of $\mathbb{R}^n$ with any desired degree of accuracy. Before presenting the theorem itself, the following definition is necessary.
\begin{defin}
    For any measurable function $g: \mathbb{R} \longrightarrow \mathbb{R}$ and any $n\in \mathbb{N}$, let $\mathcal{G}$ be the class of functions such that:
    \begin{equation}
        \mathcal{G}(g)=\left\{  f: \mathbb{R}^n \longrightarrow \mathbb{R}| \sum\limits^q_{j=1}\beta_j \prod\limits^{l_j}_{k=1} g(A_{jk}(x)), \quad x \in \mathbb{R}^n,\quad \beta_j \in \mathbb{R}, \quad A_{jk} \in \mathbb{A}^n, \quad l_j \in \mathbb{N}, \quad q=1,2,... \right\}
    \end{equation}
\end{defin}
Note that, in the definition above, $\mathbb{A}^n$ represents the set of all affine functions from $\mathbb{R}^n$ to $\mathbb{R}$. The Universal Approximation Theorem is given next.

\begin{theo}
    Let $g: \mathbb{R} \to \mathbb{R}$ be a continuous, nonconstant activation function. Then, for any compact set $K \subset \mathbb{R}^n$ and any continuous function $f \in C(K)$, the set of functions $\mathcal{G}(g)$, consisting of finite linear combinations of $g$ applied to affine transformations of $\mathbb{R}^n$, is uniformly dense in $C(K)$.
\end{theo}

The $H^1$  norm, and consequently the $H^1$ error, is widely used in numerical approximation methods for partial differential equations because it evaluates both the solution’s accuracy and its smoothness or gradient behavior. The $H^1$ error is a measure of the error between the approximated solution $u_h$ and the reference solution $u$ in the Sobolev space $H^1(\Omega)$. It is defined as
\begin{equation}
    \|u-u_h\|_{H^1(\Omega)} = \left( \|u-u_h\|^2_{L^2(\Omega)} + \| \nabla (u-u_h) \| ^2_{L^2(\Omega)}\right)^{1/2},
\end{equation}
where
\begin{equation}
    \|u-u_h\|^2_{L^2(\Omega)} = \left( \int \limits_\Omega |u(x) - u_h(x)|^2 \right)^{1/2},
\end{equation}
and
\begin{equation}
    \|\nabla(u-u_h)\|^2_{L^2(\Omega)} = \left( \int \limits_\Omega |\nabla(u(x) - u_h(x))|^2 \right)^{1/2}.
\end{equation}

The Johnson-Lindenstrauss Lemma guarantees that a high-dimensional set of points can be embedded into a lower-dimensional space while approximately preserving pairwise distances between the points.

\begin{theo}[Johnson-Lindenstrauss Lemma]
    Let $0 < \varepsilon < 1$, and let \( U \subset \mathbb{R}^d \) be a set of \( n \) points. If \( k \) is a positive integer satisfying
    \begin{equation}
        k \geq \frac{24 \log(n)}{\varepsilon^2 (3 - 2\varepsilon)},
    \end{equation}
    then there exists a mapping \( f : \mathbb{R}^d \to \mathbb{R}^k \) such that for all \( x_i, x_j \in U \),
    \begin{equation}
        (1 - \varepsilon) \|x_i - x_j\|^2 \leq \|f(x_i) - f(x_j)\|^2 \leq (1 + \varepsilon) \|x_i - x_j\|^2.
    \end{equation}
\end{theo}

As can be observed, the target dimension $k$ depends only on the number of points $n$  and the desired distortion  $\varepsilon$ , and not on the original dimension  $d$. This makes the lemma particularly valuable in dealing with very high-dimensional data.

\end{document}